# Phase4DFD: Multi-Domain Phase-Aware Attention for Deepfake Detection

Zhen-Xin Lin[1][0009-0003-6058-3957] and Shang-Kuan Chen[2][0000-0002-5879-4048]

[1,2] Department of Computer Science and Engineering, Yuan Ze University, No. 135, Yuandong Rd., Zhongli Dist., Taoyuan City 320315, Taiwan (R.O.C.)

[1]s1126070@mail.yzu.edu.tw
[2]cotachen@saturn.yzu.edu.tw

**Abstract.** Recent deepfake detection methods have increasingly explored frequency-domain representations to reveal manipulation artifacts that are difficult to detect in the spatial domain. However, most existing approaches rely primarily on spectral magnitude, implicitly underexploring the role of phase information. In this work, we propose Phase4DFD, a phase-aware frequency-domain deepfake detection framework that explicitly models phase–magnitude interactions via a learnable attention mechanism. Our approach augments standard RGB input with Fast Fourier Transform (FFT) magnitude and local binary pattern (LBP) representations to expose subtle synthesis artifacts that remain indistinguishable under spatial analysis alone. Crucially, we introduce an input-level phase-aware attention module that uses phase discontinuities commonly introduced by synthetic generation to guide the model toward frequency patterns that are most indicative of manipulation before backbone feature extraction. The attended multi-domain representation is processed by an efficient BNext-M backbone, with optional channel-spatial attention applied for semantic feature refinement. Extensive experiments on the CIFAKE and DFFD datasets demonstrate that our proposed model Phase4DFD outperforms state-of-the-art spatial and frequency-based detectors while maintaining low computational overhead. Comprehensive ablation studies further confirm that explicit phase modeling provides complementary and non-redundant information beyond magnitude-only frequency representations.

## 1. Introduction

The rapid advancement of generative models has dramatically increased the realism and accessibility of synthetic facial media. Since the introduction of generative adversarial networks (GANs) [1], increasingly powerful architectures such as StyleGAN and its variants [2, 3] and diffusion-based models [4, 5] have enabled the generation of highly photorealistic images. While these developments support a wide range of creative and practical applications, they also raise serious concerns such as digital trust, media authenticity, and societal impact [6, 7, 8, 9]. As a result, deepfake detection has emerged as a critical research problem in computer vision and pattern recognition.

Early deepfake detection methods primarily focused on spatial-domain cues, exploiting visual artifacts, texture inconsistencies, or semantic irregularities in manipulated facial regions. These approaches include head-pose inconsistency analysis [10], capsule-based architectures [11], and CNN-based detectors trained on large-scale manipulated datasets [12, 13]. Although these spatial-based methods have achieved promising performance under controlled conditions, they often struggle when manipulations are visually subtle, heavily post-processed, or generated by increasingly sophisticated models which are designed to minimize detectable artifacts [14, 15, 16].

To address the limitations of spatial-only analysis, recent studies have explored frequency-domain representations [31], motivated by the observation that generative models may introduce artifacts that are more apparent in the spectral domain than in pixel space. Prior work has demonstrated that discrepancies in Fourier spectra and frequency statistics can reveal traces of synthetic image generation [17, 18], leading to frequency-based deepfake detection methods that outperform purely spatial approaches under certain conditions [19, 20, 21]. These findings indicate that frequency-domain analysis captures complementary cues that remain effective even when visual artifacts are highly realistic.

However, most existing frequency-based deepfake detectors rely predominantly on spectral magnitude information, either discarding phase entirely or encoding it implicitly within learned representations [17, 20, 21]. This design choice ignores a fundamental property of frequency representations while magnitude captures the global energy distribution across frequencies, phase encodes spatial alignment and structural consistency, which is crucial for preserving natural image structure [22].

Generative models preserve magnitude statistics to enhance visual realism while introducing subtle phase inconsistencies due to imperfect spatial synthesis, blending operations, or upsampling artifacts. Such inconsistencies are difficult to capture using magnitude-only frequency features or spatial-domain representations alone. Motivated by this observation, we argue that phase information constitutes a complementary and underutilized cue for deepfake detection.

In this work, we propose Phase4DFD, a phase-aware frequency-domain deepfake detection framework that directly models phase-magnitude interactions through a lightweight attention mechanism. By adaptively emphasizing manipulation-sensitive frequency components, the proposed approach captures subtle generative artifacts that are underexplored by existing magnitude-focused frequency detectors. Experiments conducted on the CIFAKE [34] and the DFFD datasets [12] demonstrate that explicit phase modeling provides consistent performance improvements over spatial-only and magnitude-based frequency approaches while maintaining low computational overhead.

The main contributions of our work are summarized as follows:

1. An input-level, phase-aware attention mechanism is introduced to explicitly model phase–magnitude interactions in the frequency domain. By exploiting phase discontinuities intrinsic to synthetic image generation, this module guides feature extraction toward manipulation-sensitive cues prior to spatial processing.

2. A unified representation that integrates RGB appearance, FFT magnitude, and Local Binary Pattern (LBP) features is developed to provide complementary spatial, spectral, and textural information for robust deepfake detection.
3. Incorporating phase-aware frequency modeling into a lightweight BNext-M backbone enables strong detection performance with minimal computational overhead, making the proposed approach suitable for practical deployment.

## 2. Related Work

### 2.1. Deepfake Generation and Its Impact

Recent advances in deepfake generation have significantly increased the realism and availability of manipulated facial media. Since the introduction of generative adversarial networks (GANs) [1], generator architectures such as StyleGAN and its variants [2, 3] enabled high-quality facial synthesis with fine control over appearance. More recently, diffusion-based models [4, 5] further improved image realism by reducing visible artifacts. The widespread accessibility of these generative models has raised serious concerns about media authenticity, misinformation, and public trust [6, 7, 8, 9], which has driven growing interest in reliable deepfake detection methods.

### 2.2. Spatial-Domain Deepfake Detection

Early deepfake detection approaches primarily focused on spatial-domain cues, including inconsistencies in facial geometry, texture statistics, and semantic artifacts. These methods analyze abnormal head pose relationships [10], apply capsule networks to model part-whole relationships [11], or train convolutional neural networks on large-scale manipulated datasets [12, 13]. Later works explored local texture descriptors [29], visual artifact analysis [30], and self-supervised or data augmentation strategies to improve generalization [15, 16]. Despite strong performance in controlled settings, spatial-domain detectors often struggle when manipulations are visually subtle, heavily post-processed, or produced by advanced generation models that suppress visible artifacts [14].

### 2.3. Frequency-Domain Deepfake Detection

To overcome the limitations of spatial-only analysis, recent studies have investigated frequency-domain representations for deepfake detection. Several works show that synthetic images exhibit distinctive discrepancies in Fourier spectra that can be used for detection [17, 18]. Building on this observation, frequency-based methods analyze spectral magnitude statistics or frequency biases to distinguish real and fake images [19, 20, 21]. Other studies demonstrate that artifacts introduced by upsampling operations can also be detected in the frequency domain [25]. These results suggest that frequency-domain analysis provides complementary cues that are less dependent on visual realism. However, most existing frequency-based detectors focus primarily on spectral magnitude, while phase information is either discarded or only implicitly encoded [18, 20, 21]. As a result, potentially useful phase cues remain underexplored in current frequency-based deepfake detection frameworks.

### 2.4. Phase Information and Attention Mechanisms

The importance of phase information for preserving spatial structure and perceptual quality has long been established in signal processing [22]. In computer vision, phase-related representations have been shown to capture structural information that is complementary to magnitude-based features [23]. Despite these insights, explicit phase modeling has received limited attention in deepfake detection research.

Attention mechanisms have been widely used to emphasize manipulation-sensitive features in forgery detection and related vision tasks. Channel-wise and spatial attention modules, such as CBAM [32], as well as multi-attentional deepfake detectors [19], demonstrate that adaptive feature reweighting can improve detection performance by emphasizing manipulation-sensitive regions.

### 2.5. Positioning of Our Work

In contrast to existing approaches, we incorporate phase–magnitude interactions in the frequency domain using a lightweight attention mechanism. Rather than treating phase information as secondary, the proposed framework utilizes phase cues as a complementary signal for detecting subtle generative artifacts. By combining classical insights from signal processing with modern deepfake detection techniques, the proposed method extends magnitude-focused frequency approaches in a principled and effective manner.

## 3. Proposed Method

### 3.1. Overall Architecture

Figure 1 illustrates the overall architecture of Phase4DFD. Given an input image, the framework first constructs an augmented input representation by combining RGB appearance, frequency-domain magnitude, and local texture cues. An input-level phase-aware attention mechanism then highlights frequency components associated with abnormal phase–magnitude relationships. The attended representation is projected back to RGB space and processed by a lightweight BNext M backbone. Finally, deep features may optionally be refined using a standard channel–spatial attention module before classification. This design separates frequency-guided input modulation from feature-level refinement, allowing the contribution of each component to be examined independently.

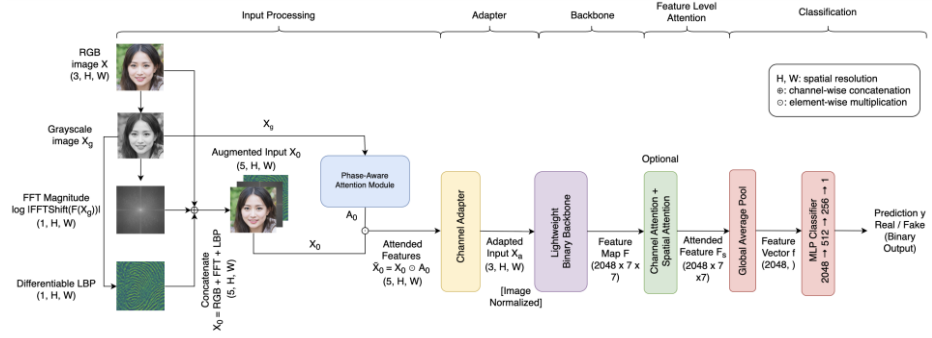

**Fig. 1.** Overall Architecture of Phase4DFD framework. RGB input is augmented with FFT magnitude and LBP to form a 5-channel tensor. Phase-aware input attention computes phase from grayscale RGB and produces $A_0$ to reweight the augmented input (⊙). A 1×1 adapter maps 5→3 channels for the BNext-M backbone, followed by optional channel-spatial attention and classification.

### 3.2. Input Processing and Augmented Representation

Given an input RGB image $X \in R^{3 \times H \times W}$, it is first converted to a grayscale image and frequency-domain magnitude spectrum is computed using a two-dimensional Fourier transform:

$$M = \log \left| FFTShift\left(F\left(X_g\right)\right)\right| \tag{1}$$

where $X_g$ denotes the grayscale image computed from RGB channels, $F(.)$ represents 2D Fourier transform, FFTshift (.) centers the zero-frequency component, and the logarithm improves numerical stability and dynamic range. A differentiable Local Binary Pattern (LBP) map is extracted from $X_g$ to encode local texture transitions that are sensitive to subtle synthesis artifacts [29]. The RGB channels, FFT magnitude, and LBP map are concatenated to form an augmented input tensor:

$$X_0 \in \mathbb{R}^{5 \times H \times W} \tag{2}$$

### 3.3. Phase-Aware Input-Level Attention

To model phase-magnitude inconsistencies introduced by generative models, an input-level phase-aware attention mechanism is applied, as shown in Figure 2. Unlike feature-level attention, this module operates directly on the augmented input $X_0$. Phase information is computed from the grayscale representation derived from the original RGB image. Specifically, the phase spectrum is obtained as:

$$\Phi = \angle\left(FFTshif\left(F\left(X_g\right)\right)\right) \tag{3}$$

where $\angle(\cdot)$ denotes the phase operator and normalization [0, 1] ensures numerical compatibility with convolutional layers. Magnitude and phase cues are processed through separate convolutional pathways and fused to generate an attention map:

$$A_0 \in \mathbb{R}^{5 \times H \times W} \tag{4}$$

The attended input is obtained via element-wise modulation:

$$\widetilde{X_0} = X_0 \odot A_0 \tag{5}$$

where $\odot$ denotes element-wise multiplication, and $A_0$ adaptively reweights augmented input channels.

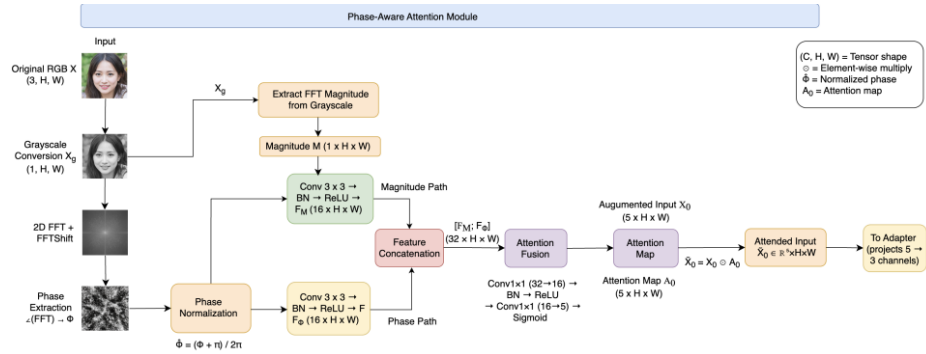

**Fig. 2.** Phase-aware input attention module. Phase is computed from the grayscale version of the original RGB image via FFT and normalized to [0,1]. FFT magnitude (from the augmented input) and normalized phase are processed by two convolutional branches, fused to generate $A_0$, and applied to the 5-channel input by element-wise multiplication ($\odot$).

### 3.4. Channel Adapter and Backbone Network

Since the backbone expects a three-channel input, a 1 x 1 convolutional channel adapter projects the attended tensor from five channels to three channels:

$$X_a \in \mathbb{R}^{3 \times H \times W}$$

The adapted input is then fed into a BNext-M backbone [30], which follows a hierarchical convolutional design to extract deep spatial features [31, 32]. The backbone remains unchanged across all experimental settings to ensure that performance differences arise from input modeling rather than backbone capacity.

### 3.5. Feature-Level Channel-Spatial Attention (optional)

Optionally, a feature-level channel-spatial attention mechanism is applied to the backbone output feature map, which is depicted in Figure 3:

$$F \in \mathbb{R}^{2048 \times 7 \times 7} \tag{7}$$

Following the CBAM formulation [32], channel attention recalibrates inter-channel dependencies to produce $A_c$, followed by spatial attention that highlights informative spatial locations via $A_s$. The refined feature map is obtained as:

$$F_s = (F \odot A_c) \odot A_s \qquad (8)$$

where $A_c$ and $A_s$ denote channel and spatial attention maps, respectively. This attention mechanism serves as a standard feature refinement step and is not intended to capture frequency or phase information. As shown in Section 4.5, its impact is limited compared to the proposed input-level phase-aware modulation.

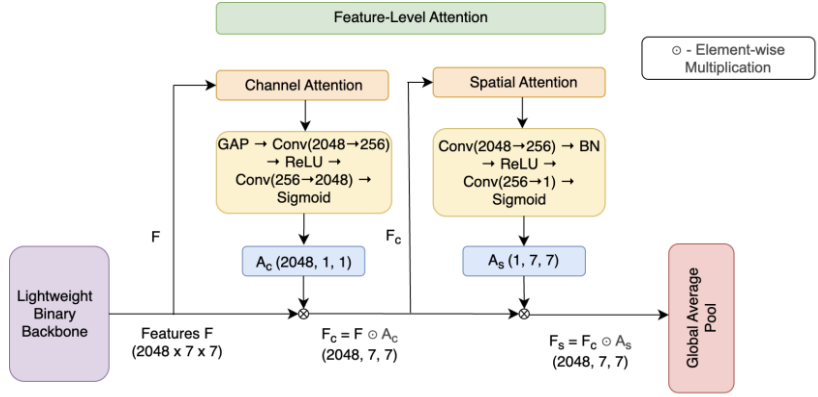

**Fig. 3.** Feature-level channel-spatial attention.

### 3.6. Classification

The refined feature map is globally averaged to produce a compact feature vector, which is fed into a multi-layer perceptron classifier to predict whether the input image is real or manipulated. Overall, the proposed framework prioritizes input-level phase-aware frequency modulation, enabling the backbone to extract features from representations already biased toward manipulation-related cues. Feature-level attention is treated as an auxiliary refinement rather than a core component, which aligns with the empirical observations reported in Section 4.

## 4. Experiments

### 4.1. Datasets

The proposed Phase4DFD framework is evaluated on two widely used deepfake image benchmarks: CIFAKE [34] and DFFD [12], which represent complementary synthetic generation settings.

CIFAKE [34] is a large-scale synthetic extension of CIFAR-10 [37], consisting of 120,000 images evenly split between real samples and fake images generated using Stable Diffusion [4]. The dataset follows the standard split of 100,000 training images and 20,000 test images. CIFAKE primarily evaluates robustness to diffusion-based image synthesis under low-resolution conditions.

DFFD [12] is a large-scale face forgery dataset covering multiple manipulation types, including identity swapping, expression transfer, attribute editing, and fully synthetic faces. It contains approximately 58k real images and 240k fake images generated using PGGAN and StyleGAN architectures [2]. The dataset is partitioned into 50% training, 5% validation, and 45% testing subsets, providing a challenging evaluation setting with diverse forgery patterns.

### 4.2. Implementation Details

All experiments are implemented using PyTorch 2.0 and PyTorch Lightning 2.1. Training and evaluation are conducted on a single NVIDIA RTX 3090 GPU (24 GB) with CUDA 12.1.

### 4.3. Training Strategy

**Data Preprocessing and Augmentation.** For CIFAKE, images are resized to 224×224, while DFFD images are resized to 192×192 to balance resolution and computational cost. Data augmentation during training includes random horizontal flipping (probability 0.5), random rotation within ±15°, color jittering (brightness, contrast, saturation set to 0.2), and random resized cropping with a scale range of 0.8-1.0. All Augmentations are applied prior to FFT magnitude and LBP extraction to ensure consistency between spatial and frequency-domain representations. After the channel adapter, standard ImageNet normalization (μ = [0.485, 0.456, 0.406], σ = [0.229, 0.224, 0.225]) is applied to match the pretrained backbone statistics.

**Loss Function.** The model is optimized using a weighted combination of Binary Cross-Entropy (BCE) and Focal Loss [38]:

$$L_{train} = 0.7 L_{BCE} + 0.3 L_{Focal} \quad (9)$$

The BCE term incorporates a positive class weighting

$$w_{pos} = N_{real}/N_{fake},$$

to mitigate class imbalance. Focal Loss utilizes a focusing parameter $\gamma = 2$ to emphasize hard-to-classify samples and improve robustness to visually subtle manipulations.

**Optimization and Training Strategy.** Optimization is performed using AdamW [38], with a cosine annealing learning rate scheduler [39] applied throughout training. A dataset-specific two-stage training strategy is adopted to stabilize optimization while preserving pretrained knowledge.

*CIFAKE training.* The BNext-M backbone is frozen for the first 5 epochs, during which only the input adapter, attention modules, and classifier are trained with a learning rate of $10^{-3}$. The backbone is then unfrozen and jointly trained for an additional 15 epochs, using a reduced learning rate of $10^{-4}$ for backbone parameters while maintaining $10^{-3}$ for newly introduced components.

*DFFD training.* Due to the larger scale and higher diversity of manipulations, the backbone remains frozen for the first 10 epochs, followed by 15 epochs of joint fine-tuning using the same learning rate configuration as CIFAKE.

### 4.4. Evaluation Metrics

Performance is assessed using standard binary classification metrics, including Accuracy, Area Under the ROC Curve (AUC), Precision, Recall, and F1-score. For ablation studies, Accuracy and AUC on the DFFD dataset [12] are reported to quantify the individual contributions of phase-aware attention, frequency-domain inputs, and feature-level refinement components.

### 4.4. Experimental Results

**Performance on DFFD.** Table 1 reports the detection performance on the DFFD dataset in terms of accuracy and AUC, with all methods evaluated under their standard settings. Phase4DFD achieves the best overall performance, reaching 99.46% accuracy and 99.95 AUC, surpassing both conventional CNN-based detectors and recent BNext variants. Compared to the BNext-M unfrozen backbone, Phase4DFD improves accuracy by +0.71% while using the same backbone capacity, indicating that the performance gains are attributable to the proposed phase-aware frequency modeling rather than architectural scaling. Comparison with lightweight BNext-T and BNext-S models, Phase4DFD consistently delivers superior performance, demonstrating that integrating phase-aware input modulation provides additional discriminative power beyond backbone depth or width.

**Table 1.** Detection performance on the DFFD dataset. Results are reported in terms of accuracy and AUC. **Bold** and underlined values indicate the best and second-best results, respectively. All methods are evaluated under their standard settings. Improvements achieved by Phase4DFD are obtained without increasing backbone capacity, highlighting the benefit of the proposed phase-aware frequency modeling rather than architectural scaling.

|   | Method | Accuracy | AUC |
|---|--------|----------|-----|
| [12] | Xception | - | 99.64 |
|   | VGG16 | - | 99.67 |
| [31] | BNext -T | 98.95 | <u>99.94</u> |
|   | BNext -S | <u>99.01</u> | <u>99.94</u> |
|  | BNext - M | 98.75 | 99.92 |
| Ours | Phase4DFD | **99.46** | **99.95** |

**Performance on CIFAKE.** Table 2 presents the results of the CIFAKE test set. Our Phase4DFD achieves 98.62% accuracy and 99.88 AUC, exceeding all baseline CNN architectures and improving upon the BNext-M baseline by +1.27% accuracy. These improvements are obtained without increasing the number of parameters, confirming that the proposed approach enhances detection capability through input-level modeling rather than increased network capacity. The performance gap is particularly evident when compared to standard CNN and ResNet-based detectors, highlighting the benefit of incorporating frequency and phase-aware representations when handling diffusion-generated images.

**Table 2.** Performance comparison on the CIFAKE test set, including accuracy and AUC. **Bold** and underlined values denote the best and second-best results, respectively.

|  | Method | Accuracy | AUC |
|---|---|---|---|
| [40] | CNN | 86.00 | 93.00 |
|  | ResNet | 95.00 | 99.00 |
|  | VGGNet | 96.00 | 99.00 |
|  | DenseNet | <u>98.00</u> | 99.00 |
| [31] | BNext-T | 97.29 | <u>99.65</u> |
|  | BNext-S | 96.96 | 99.55 |
|  | BNext-M | 97.35 | 99.62 |
| Ours | Phase4DFD | **98.62** | **99.88** |

**Class-wise PerformanceAnalysis.** To further assess prediction balance, Table 3 reports the class-wise precision, recall, and F1-score on CIFAKE. Phase4DFD demonstrates well-balanced performance across both real and fake classes, achieving comparable F1-scores (98.62) for fake and real images. This balance indicates robustness to both false positives and false negatives that the proposed model avoids bias toward a single class.

**Table 3.** Class-wise precision, recall, and F1-score for fake and real samples on the CIFAKE dataset. The proposed method shows balanced performance across both classes, indicating robustness to false positives and false negatives and avoiding bias toward a single class.

|  | | Fake | | | Real | | |
|---|---|---|---|---|---|---|---|
|  | Method | Precision | Recall | F1-Score | Precision | Recall | F1-Score |
| [40] | CNN | 86.00 | 87.00 | 87.00 | 87.00 | 85.00 | 86.00 |
|  | ResNet | **99.00** | 91.00 | 95.00 | 91.00 | **99.00** | 95.00 |
|  | VGGNet | 97.00 | 95.00 | 96.00 | 95.00 | 97.00 | 96.00 |
|  | DenseNet | 98.00 | <u>98.00</u> | <u>98.00</u> | <u>98.00</u> | 98.00 | <u>98.00</u> |
| Ours | Phase4DFD | <u>98.24</u> | **98.83** | **98.62** | **98.83** | <u>98.41</u> | **98.62** |

Across both datasets, Phase4DFD demonstrates consistent improvements over spatial-only and frequency-magnitude based baselines under standard evaluation protocols. The results validate that phase-aware frequency modeling provides complementary discriminative cues, enabling improved deepfake detection performance without increasing backbone complexity.

### 4.5. Ablation Study

Table 4 presents an ablation study on the DFFD dataset, analyzing the impact of individual input modalities and attention mechanisms. Starting from the RGB-only baseline, adding FFT magnitude or LBP individually results in only marginal performance changes (+0.03% and +0.01% accuracy, respectively), indicating that naïve incorporation of frequency or texture cues provides limited benefit when used in isolation. Combining FFT magnitude and LBP without dedicated modeling further degrades performance, suggesting that simple feature concatenation may introduce redundancy or noise rather than complementary information.

Applying generic feature-level channel-spatial attention also fails to improve performance, and in some cases slightly reduces accuracy. This observation implies that standard attention mechanisms, when applied after backbone feature extraction, are insufficient for capturing the subtle artifacts introduced by modern generative models. In contrast, introducing the proposed phase-aware input-level attention leads to the largest and most consistent performance improvement, achieving 99.46% accuracy, a gain of +0.23% over the RGB baseline. Notably, this improvement is achieved without additional backbone capacity or feature-level refinement. The result highlights that explicitly modeling phase-magnitude relationships at the input level is more effective than either feature fusion or generic attention applied at later stages. These findings support the design choice of prioritizing phase-aware input modulation as the core component of Phase4DFD.

**Table 4.** Ablation study on the DFFD dataset evaluating the impact of individual input modalities and attention mechanisms. Results highlight that feature fusion or generic feature-level attention provides limited benefit, whereas the proposed phase-aware input-level attention consistently yields the largest performance improvement.

| Model Variant | Input/ Module | Accuracy (%) |
| --- | --- | --- |
| Baseline | RGB | 99.23 |
| Baseline + Added Features | RGB + FFT Magnitude | <u>99.26</u> |
| | RGB + LBP | 99.24 |
| | RGB + FFT + LBP | 99.13 |
| | RGB+ FFT + LBP + Channel & Spatial Attention | 99.18 |
| | RGB + FFT+ LBP + Phase-Aware + Channel & Spatial Attention | 99.11 |
| | RGB + FFT + LBP + Phase-Aware Attention | **99.46** |

## 5. Conclusion

In this study, we proposed Phase4DFD, a phase-aware frequency-domain framework for deepfake image detection which incorporates phase information into the detection pipeline through input-level attention. By augmenting RGB images with FFT magnitude and local texture cues and prioritizing phase-magnitude interactions before backbone processing, the proposed method enables the network to focus on manipulation-sensitive spectral patterns that are often overlooked by spatial or magnitude-only approaches.

Extensive experiments on the CIFAKE and DFFD benchmarks reveal that Phase4DFD consistently outperforms strong CNN and BNN baselines under standard evaluation settings, while maintaining identical backbone capacity and low computational overhead. Ablation studies further confirm that the feature fusion or generic feature-level attention provides limited benefit, whereas the proposed phase-aware input modulation yields the most significant performance gains. These results highlight the importance of modeling phase information as a complementary cue for detecting increasingly realistic synthetic images.

Overall, this work shows that revisiting classical signal properties, such as phase, and integrating them into modern deepfake detection architectures offers an effective and efficient path forward. Future work will explore extending the proposed framework to cross-dataset generalization and robustness against emerging generation techniques.